\documentclass[journal]{IEEEtran}

\usepackage[utf8]{inputenc}
\usepackage[T1]{fontenc}
\usepackage{microtype}
\usepackage{booktabs}
\usepackage{array}
\usepackage{xcolor}
\usepackage{amsmath}
\usepackage{amssymb}

\usepackage{amsthm}
\usepackage{graphicx}
\usepackage{tabularx}
\usepackage{float}
\usepackage[hidelinks]{hyperref}
\usepackage{xurl}

\newtheorem{definition}{Definition}
\newcommand{\zenododoi}{\href{https://doi.org/10.5281/zenodo.23724735}{10.5281/zenodo.23724735}}

\newcommand{\DGeometry}{2.01}
\newcommand{\DQOne}{0.215}

\newcommand{\DQFour}{0.588}
\newcommand{\DQfourKone}{0.00}
\newcommand{\DQfourKten}{0.48}
\newcommand{\DQfourKthirty}{0.588}
\newcommand{\DlevelFive}{1.34}

\newcommand{\DmaxQFour}{0.827}

\newcommand{\PosQFour}{38}
\newcommand{\RQOne}{96.67}

\newcommand{\RQFour}{97.40}
\newcommand{\SdoMaxQOne}{98.0}

\newcommand{\SdoMaxQFour}{98.6}

\newcommand{\SdoMeanQFour}{97.98}
\newcommand{\SdoMinQOne}{95.2}

\newcommand{\SdoMinQFour}{97.4}
\newcommand{\SelQOne}{0.58}

\newcommand{\SelQFour}{1.27}

\newcommand{\UQTwo}{97.70}

\newcommand{\UQFour}{97.99}

\newcommand{\accLevelFiveNano}{91.5}
\newcommand{\accLevelFiveSeventy}{54.2}
\newcommand{\accLevelFiveSuper}{92.3}
\newcommand{\accLevelOneSeventy}{96.4}
\newcommand{\accNano}{96.1}
\newcommand{\accSeventy}{76.2}

\newcommand{\aurocNano}{0.89}

\newcommand{\aurocSuper}{0.88}
\newcommand{\bestSingle}{96.13}
\newcommand{\cappedNano}{9}
\newcommand{\cappedSeventy}{6}
\newcommand{\cappedSuper}{16}
\newcommand{\cappedTotal}{31}
\newcommand{\distinctNano}{1.54}
\newcommand{\distinctSeventy}{3.16}
\newcommand{\distinctSuper}{1.37}
\newcommand{\eightEarly}{4,354}
\newcommand{\eightFinal}{4,565}
\newcommand{\eightFinalHours}{3.0}
\newcommand{\eightPreCapped}{2}
\newcommand{\eightPreN}{8}
\newcommand{\gpuMinPerKNano}{5.4}

\newcommand{\gpuMinPerKSuper}{25.5}
\newcommand{\gpuRatioNanoSuper}{4.7}
\newcommand{\hAllPerfect}{243}

\newcommand{\hNanoGtSuper}{41}
\newcommand{\hSeventyGtSuper}{20}
\newcommand{\hSuperGtNano}{46}
\newcommand{\hSuperGtSeventy}{231}
\newcommand{\headMinSeventy}{31}

\newcommand{\lostThirtyTwoNano}{2}

\newcommand{\lostThirtyTwoSuper}{5}
\newcommand{\medCorrectNano}{1,833}

\newcommand{\medCorrectSuper}{2,184}
\newcommand{\medWrongNano}{9,924}

\newcommand{\medWrongSuper}{12,265}
\newcommand{\mixedNano}{65}
\newcommand{\mixedSeventy}{195}
\newcommand{\mixedSuper}{63}

\newcommand{\passThirtySeventy}{91.0}

\newcommand{\passThreeNano}{97.99}

\newcommand{\passTwoNano}{97.54}

\newcommand{\periodicTotal}{25}
\newcommand{\sqPeriodicNano}{10.2}
\newcommand{\sqPeriodicSeventy}{30.1}
\newcommand{\sqPeriodicSuper}{11.2}
\newcommand{\sqShareNano}{11.4}
\newcommand{\sqShareSeventy}{60.2}
\newcommand{\sqShareSuper}{12.8}
\newcommand{\tailMinSeventy}{39.65}
\newcommand{\tailShareSeventy}{53.9}

\newcommand{\tokPninetynineNano}{18,096}
\newcommand{\tokPninetynineSeventy}{2,039}
\newcommand{\tokPninetynineSuper}{22,804}
\newcommand{\tokTotalNano}{46.0}
\newcommand{\tokTotalSeventy}{9.5}
\newcommand{\tokTotalSuper}{57.3}
\newcommand{\tpsNano}{9,806}
\newcommand{\tpsSeventy}{2,147}
\newcommand{\tpsSuper}{2,597}
\newcommand{\wallFullSeventy}{73.6}
\newcommand{\wallNano}{78}
\newcommand{\wallSeventy}{74}
\newcommand{\wallShareNanoSuper}{21}
\newcommand{\wallSuper}{368}

\newcommand{\wilNanoSuper}{+0.05}

\title{When Strong Reasoners Join the Pool:\\Small Union Premiums and Costly Runaway\\Generations in LLM Routing}

\author{Teng-Ruei~Chen%
\thanks{T.-R. Chen is with the Institute of Bioinformatics and Systems Biology, National Yang Ming Chiao Tung University, Hsinchu, Taiwan, and also with Krixvon, Taipei, Taiwan (e-mail: ymchen.bi04g@g2.nctu.edu.tw; ORCID: 0000-0001-8995-334X).}%
\thanks{Technical report, November 2026. This report is a companion to the author's manuscript on routing-oracle estimands~\cite{chen2026whatoracle}: it applies that framework to current reasoning models on new hardware and adds no result to that manuscript. Every reported number, table and figure is generated by one script from the frozen run artifacts.}}

\begin{document}
\bstctlcite{IEEEexample:BSTcontrol}
\markboth{Technical Report, November 2026}{Technical Report, November 2026}
\maketitle

\begin{abstract}
Routing benchmarks often judge a router against a hindsight oracle that counts a
query as solved whenever any model's recorded answer is correct, but under
stochastic decoding that oracle mixes two different objects: the clairvoyant
ceiling of committing to one model per query and the premium of calling several
models and keeping any correct answer. A companion study measured this premium at
$3.5$ percentage points (pp) for 11 small open models on MATH-500; how large it is once
current reasoning models enter the pool was unknown. We measure it with a
fixed-protocol, pre-committed generation on a single NVIDIA B300: all
$500$ MATH-500 problems, $30$ seeded samples per problem, Llama-3.3-70B-Instruct
and NVIDIA's Nemotron Nano 9B v2 and Nemotron Super 49B v1.5 (a NAS derivative of
the same 70B) under vendor sampling, a 65{,}536-token cap, a pinned Math-Verify
scorer, and an append-only attempt ledger, for $45{,}000$ scored responses. The two
reasoning models both reach \accNano\% accuracy against \accSeventy\% for the 70B
baseline. The empirical product-coupling premium is \DQOne--\DQFour\,pp across
all two- and three-model pools; the empirical plug-in selection headroom over always
calling Nano is \SelQFour\,pp; and resampling Nano three times gives practically the
same empirical any-correct rate as calling all three models once. Only \cappedTotal\
responses reached the cap, \periodicTotal\ of them with exactly periodic tails, yet the
capped responses account for up to \sqShareSeventy\% of a model's output-length-squared
proxy, and long generations made a fourth planned model, Llama-3.1-8B, too slow to
finish. The premium is far below the companion's $3.5$\,pp, but models, pool size and
protocol changed together; an oracle gap should therefore be reported with the pool,
the protocol, the samples per cell and the decoding budget that produced it.
\end{abstract}


% ------------------------------------------------------------------------------
\section{Introduction}
\begin{figure*}[t]\centering
\includegraphics[width=\textwidth]{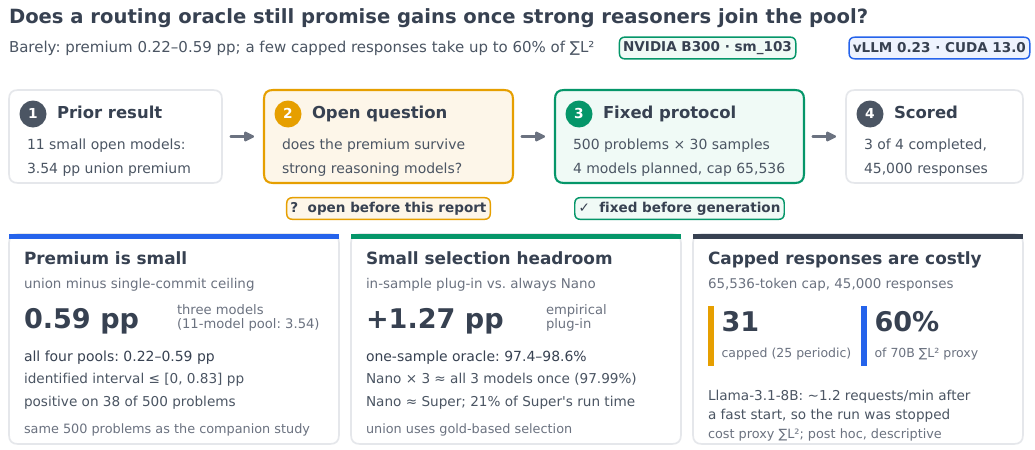}
\caption{Overview.  \emph{Top:} the question; the prior result of the companion
study~\cite{chen2026whatoracle}, a $3.54$\,pp union premium for 11 small open models on
MATH-500 under its own protocol; and the protocol fixed before generation on one NVIDIA
B300, planned for four models, of which three completed (the scoring amendment is
described in Section~\ref{sec:protocol}).  \emph{Bottom:} the findings.  The empirical
union premium over the single-commit ceiling is \DQOne--\DQFour\,pp across pools.  The
empirical plug-in selection headroom over always calling Nano is \SelQFour\,pp, an oracle
built from one sample per model ranges over \SdoMinQFour--\SdoMaxQFour\%, and resampling
Nano three times gives practically the same empirical any-correct rate as calling all three
models once.  Of $45{,}000$ responses, \cappedTotal\ hit the cap, \periodicTotal\ of them
with exactly periodic tails; the capped responses take \sqShareSeventy\% of the 70B
output-length-squared proxy, and the Llama-3.1-8B run was stopped for cost.}
\label{fig:overview}
\end{figure*}
Serving large language models (LLMs) is a cost--quality problem,
and routing---choosing one model per query from a pool---is one of its most studied
answers~\cite{chen2023frugalgpt,ong2024routellm,hu2024routerbench}. Routing
benchmarks commonly compare a router with an \emph{oracle} that credits a query
whenever any pool member answered it correctly~\cite{hu2024routerbench}, and a large
router-to-oracle gap is easily read as headroom that better routing could capture.

That reading assumes the oracle is a property of the queries and the models. Under
stochastic decoding it is not: each recorded answer is one sample, so the oracle is
a union over separate samples, and it can exceed anything a router that commits
to a single model per query could attain~\cite{chen2026whatoracle}.
Our companion study~\cite{chen2026whatoracle} formalized this as two decision
classes: the clairvoyant single-commit ceiling $R_i=\max_m p_{im}$ and a named
product-coupling union $U_i^{\perp}=1-\prod_m(1-p_{im})$, whose difference
$\Delta_i^{\perp}=U_i^{\perp}-R_i$ is a premium that requires more than one call per
query. On a pool of 11 small open models it measured premiums of $0.37$, $3.54$
and $5.10$\,pp on GSM8K, MATH-500 and GPQA-Diamond~\cite{chen2026whatoracle}.

Two questions were left open. The first is how large the premium remains once current
reasoning models are in the pool: they solve most competition problems in most
samples, and a pool containing one may leave little for a union to add. The second is
practical: reasoning models write long, variable-length answers, and the decoding
budget that a fair comparison requires has a cost that small-model studies rarely
faced. Both matter to anyone who sizes a
multi-model deployment from a benchmark's oracle.

This report addresses both with one fixed-protocol measurement on a single NVIDIA
B300. We generated $30$ seeded samples for every MATH-500 problem from three
current models---Llama-3.3-70B-Instruct~\cite{grattafiori2024llama3}, NVIDIA
Nemotron Nano 9B v2~\cite{nvidia2025nano2}, and Nemotron Super 49B
v1.5~\cite{bercovich2025llamanemotron}, a neural-architecture-search derivative of
the same 70B model~\cite{bercovich2024puzzle}---under a protocol fixed before
generation, and scored all $45{,}000$ responses with a pinned verifier. The study is a
prospective empirical extension of the companion framework, not a replication of its
numbers: pinning revisions, fixing the pool before any output and drawing every
response with its own seed address two limitations the companion lists for its archived
tensors---unpinned revisions and outcome-informed pool selection---and replace
independence imposed by a resampling rule with independently seeded generation. The
contributions are:
\begin{enumerate}
\item \textbf{A measurement of the union premium with strong reasoners in the
pool.} The empirical product-coupling premium is \DQOne--\DQFour\,pp across all two-
and three-model pools, and its marginally identified interval never exceeds
$[0,\DmaxQFour]$\,pp. Under one protocol, changing only the pool moves it by a factor of
almost three; the companion's 11-model pool gave $3.54$\,pp on the same problems under
a different protocol (Section~\ref{sec:results}).
\item \textbf{A decomposition of the oracle gap that separates routing from
sampling.} Over the best single model, the empirical plug-in selection headroom is
\SelQFour\,pp and the product union adds a further \DQFour\,pp; a benchmark oracle built
from one sample per model would report anywhere between \SdoMinQFour\% and
\SdoMaxQFour\% depending only on which sample it kept; and resampling the cheapest
strong model three times gives practically the same empirical any-correct rate as
calling all three models once (Sections~\ref{sec:results} and~\ref{sec:resample}).
\item \textbf{A cost account of the decoding budget.} A uniform 65{,}536-token cap
truncated \cappedTotal\ of $45{,}000$ responses, \periodicTotal\ with exactly periodic
tails, yet the capped responses take up to \sqShareSeventy\% of a model's
output-length-squared proxy, and the 70B run spent \tailShareSeventy\% of its formal
wall time on its last 11 requests; we also report how length separates wrong from
correct answers and what tighter caps would have cut (Section~\ref{sec:runaway}).
\item \textbf{An auditable execution record.} Pre-commitment, seed manifests, an
append-only attempt ledger, a gold self-equivalence check, two-launch engine
probes and hardware and software provenance, with every deviation documented and
two archive gaps stated (Section~\ref{sec:protocol}).
\end{enumerate}
Fig.~\ref{fig:overview} summarizes the study. Section~\ref{sec:related} positions it, Section~\ref{sec:framework} recalls
the estimands, Section~\ref{sec:protocol} describes the protocol and platform,
Sections~\ref{sec:results}--\ref{sec:runaway} report results,
Section~\ref{sec:discussion} discusses implications, Section~\ref{sec:limitations} states
limitations, and
Section~\ref{sec:conclusion} concludes.

% ------------------------------------------------------------------------------
\section{Related Work}\label{sec:related}
We group prior work into three lines and compare them with this study in
Table~\ref{tab:related}.

\textbf{Routing and its oracle.} Cascades such as FrugalGPT~\cite{chen2023frugalgpt}
(2023) and learned routers such as RouteLLM~\cite{ong2024routellm} (2024) select
among models per query; RouterBench~\cite{hu2024routerbench} (2024) standardized the
evaluation and its per-query oracle; RouteLLM mainly reports cost--quality trade-offs
against a strong-model baseline, and RouterBench also uses graded quality scores. Where
an evaluation records one binary correctness label per query and model, its oracle is
a union of single samples. Our companion study~\cite{chen2026whatoracle} re-estimated
per-cell success from $30$ archived samples and formalized the single-commit ceiling,
the product-coupling union and the coupling-aware premium interval used here.

\textbf{Repeated sampling and test-time compute.} Self-consistency~\cite{wang2023selfconsistency}
(2023), the unbiased pass@$k$ estimator~\cite{chen2021codex} (2021), repeated-sampling
scaling~\cite{brown2024monkeys} (2024) and compute-optimal test-time
scaling~\cite{snell2024scaling} (2024) show that drawing more samples from one model
can rival larger models. Our RoR study~\cite{chen2026ror} (2026) asked whether a
fixed call budget is better spent resampling one model or rerouting across models;
Section~\ref{sec:resample} gives a direct instance of that comparison.

\textbf{Length, overthinking and degeneration.} Neural text degeneration into
repetition is a long-known failure of sampling~\cite{holtzman2020degeneration}
(2020); for reasoning models, overthinking---long outputs without accuracy
gains---is now well documented~\cite{chen2024overthinking,sui2025stopoverthinking}
(2024--2025). We contribute a cost account of the few responses that do not terminate
within a fixed cap.

\begin{table*}[t]
\centering\footnotesize
\caption{Positioning against prior evaluations. \checkmark: addressed;
--: not addressed.}
\label{tab:related}
\begin{tabular}{@{}lccccc@{}}
\toprule
Study & Multi-sample cell & Ceiling vs.\ union & Protocol fixed & Current & Decoding-cost \\
 & success rates & separated & before generation & reasoning models & accounting \\
\midrule
RouterBench~\cite{hu2024routerbench} & -- & -- & -- & -- & -- \\
RouteLLM~\cite{ong2024routellm} & -- & -- & -- & -- & -- \\
Repeated sampling~\cite{brown2024monkeys} & \checkmark & -- & -- & -- & \checkmark \\
Routing oracle~\cite{chen2026whatoracle} & \checkmark & \checkmark & archived tensors & -- & -- \\
\textbf{This report} & \checkmark & \checkmark & \checkmark & \checkmark & \checkmark \\
\bottomrule
\end{tabular}
\end{table*}

% ------------------------------------------------------------------------------
\section{Estimands}\label{sec:framework}
We follow the notation of the companion study~\cite{chen2026whatoracle}. Fix a
protocol: queries, model pool $P$, prompt template, decoder settings, length cap and
scorer. For query $i$ and model $m$, $p_{im}$ is the probability that one sampled
response is scored correct.

\begin{definition}[Ceiling, union and premium]
The clairvoyant single-commit ceiling is $R_i=\max_{m\in P}p_{im}$; the named
product-coupling union is $U_i^{\perp}=1-\prod_{m\in P}(1-p_{im})$; the premium is
$\Delta_i^{\perp}=U_i^{\perp}-R_i\ge0$.
\end{definition}

$R_i$ is the best a router that commits to one model per query can expect, even if
it knew every $p_{im}$; no single-commit policy exceeds it~\cite{chen2026whatoracle}.
$U_i^{\perp}$ is the success of calling every model once and keeping any correct
answer, which presupposes gold-based selection and $|P|$ calls; it is not a
deployable selector at equal cost. With only the marginals fixed, the attainable
union probabilities form the interval $[R_i,\min\{1,\sum_m p_{im}\}]$, so the premium
is only identified up to $[0,\min\{1,\sum_m p_{im}\}-R_i]$~\cite{chen2026whatoracle};
we report the product value and this interval.

We estimate $p_{im}$ by $\hat p_{im}=c_{im}/30$ from the $30$ frozen samples, plug it
into each estimand per query, and average over the $500$ queries. Three further
descriptive quantities connect the estimands to practice. The empirical plug-in
\emph{selection headroom} $\bar{\hat R}-\max_m\bar{\hat p}_m$ estimates what a clairvoyant
per-query router would add over always calling the best single model; because the best
model and the per-query maxima are chosen on the same samples, it is an in-sample
quantity. The \emph{single-draw oracle} for
sample index $j$ is $\frac{1}{500}\sum_i\max_m s_{imj}$, the oracle a benchmark would
report from one recorded sample per model. The within-model any-correct rate for
$k$ samples is estimated by the unbiased pass@$k$
$1-\binom{30-c_{im}}{k}/\binom{30}{k}$~\cite{chen2021codex}. All quantities are
descriptive for these fixed queries and this protocol; we make no population claim.

% ------------------------------------------------------------------------------
\section{Protocol and Platform}\label{sec:protocol}

\subsection{Queries, models and decoding}
The queries are all $500$ problems of MATH-500~\cite{hendrycks2021math,lightman2024verify}
(Hugging Face \texttt{HuggingFaceH4/MATH-500}, revision \texttt{2343f79f}) in source
order, each with the instruction to reason step by step and put the final answer in
\verb|\boxed{}|. No subset was drawn, so the query set is identical to the MATH-500
pool of the companion study. Each model received the prompt as a single user message
with no system message, rendered by its own pinned chat template.

Table~\ref{tab:models_cfg} lists the models and decoding. We used each vendor's
recommended sampling: temperature $0.6$ with top-$p$ $0.9$ for Llama and $0.95$ for the
Nemotron models, whose native reasoning mode stayed on. Every response is a
separate request with $n=1$ and its own seed, derived by hashing a pre-committed
salt with the model, query and sample index; the $60{,}032$ formal and preflight seeds
were checked for collisions before any generation. The completion cap was fixed at
$65{,}536$ tokens for every model; a response that reached it is scored $0$.

\begin{table}[t]
\centering\footnotesize
\caption{Models and decoding (pinned Hugging Face revisions).}
\label{tab:models_cfg}
\begin{tabular}{@{}llccc@{}}
\toprule
Model & Revision & $T$ & top-$p$ & Reasoning \\
\midrule
Llama-3.3-70B-Instruct & \texttt{6f6073b4} & $0.6$ & $0.9$ & -- \\
Nemotron Nano 9B v2 & \texttt{6533e8de} & $0.6$ & $0.95$ & on \\
Nemotron Super 49B v1.5 & \texttt{420ba7d2} & $0.6$ & $0.95$ & on \\
\midrule
\multicolumn{5}{@{}l}{All: BF16 weights and KV cache, cap $65{,}536$, 30 samples/query.}\\
\bottomrule
\end{tabular}
\end{table}

\subsection{Scoring}
The scorer extracts the last unescaped \verb|\boxed{}| of the final region---the
text after the closing \verb|</think>| for reasoning models---and compares it with
the gold answer using Math-Verify 0.9.0~\cite{mathverify2025} in a fresh,
resource-limited CPU worker per comparison, run strictly serially. Unclosed
reasoning, missing or malformed boxes and length-truncated responses score $0$.
Exceptions handled inside the verifier are recorded per response. A scorer
exception that escapes the verifier stops scoring unless it is one of three
pre-specified raising locations in the pinned source and reproduces exactly in a
fresh worker, in which case it scores $0$ and is reported; none occurred. Before
scoring, all $500$ gold answers passed a self-equivalence check in the same worker.

\subsection{Pre-commitment, ledger and amendments}
Rules, salts, query set, scorer and the confirmatory analysis were written to a dated
pre-commitment before any request. That analysis was the ceiling, product union and
premium of four-model pools, with no routing headroom. Every generation attempt was registered in an append-only,
hash-chained ledger before it started and closed with the hashes of all its files;
each model was allowed one infrastructure retry with identical configuration. Each
model's engine was launched twice before preflight to confirm that its resolved
configuration, chat template, end-of-sequence tokens and sampling parameters were
identical across launches, and an eight-request preflight per model confirmed that
every request ran with its manifest seed and that the KV cache was BF16 in the
kernels (Nano's Mamba state was float32 as required).

We record every deviation. Before any generation, the 70B model was changed from
Llama-3.1-70B to Llama-3.3-70B, the parent of Super 49B, and two engine-record
serializer defects found by the first probes were fixed; each change superseded the
commitment with an identical rule set. The plan also included Llama-3.1-8B-Instruct.
Its formal run was stopped by the author after three hours, before any response of
any model was scored: after a fast start it completed only ${\sim}1.2$ requests per
minute, so the remaining ${\sim}10{,}400$ requests would have taken several more days
(Section~\ref{sec:runaway}). The attempt is recorded as quarantined, and a dated
amendment replaced the four-model pools with the three-model pools Q1--Q4 (all pairs and
the triple) before scoring. Everything else reported here---the premium interval,
selection headroom, single-sample oracles, finite-$k$ paths, pass@$k$, paired tests, subject and level
breakdowns, and the length and loop diagnostics---is post hoc and descriptive.

Two records are not in the archive. The scorer receipt pins the gold bundle, the dataset
revision and the hash of its source file, but the bundle itself was not retrieved from
the scoring host; it can be rebuilt from the pinned revision and checked against the
recorded hashes. The superseded v3 commitment is identified by its hash in the v4
commitment, but its body is not included.

\subsection{Platform}
Table~\ref{tab:hw} summarizes the platform; the full record, including raw
\texttt{nvidia-smi} and package dumps, accompanies the data release. Each model ran
alone on the GPU through one vLLM~\cite{kwon2023vllm} engine call containing all
$15{,}000$ of its requests.

\begin{table}[t]
\centering\footnotesize
\caption{Execution platform.}
\label{tab:hw}
\begin{tabular}{@{}ll@{}}
\toprule
GPU & 1$\times$ NVIDIA B300 SXM6 AC (Blackwell) \\
 & compute capability 10.3, 148 SMs \\
GPU memory & 275{,}040 MiB HBM, ECC on \\
Power limit & 1{,}100\,W \\
Host link & PCIe Gen6 $\times$16, KVM pass-through \\
CPU / RAM & 30 vCPU AMD EPYC 9575F / 251\,GiB \\
OS / driver & Ubuntu 24.04.5, Linux 6.8 / 580.178.04 \\
CUDA stack & CUDA 13.0, PyTorch 2.11.0+cu130 \\
 & cuDNN 9.19, NCCL 2.28.9 \\
Engine & vLLM 0.23.0 (V1), FlashInfer 0.6.12 \\
Scorer & Math-Verify 0.9.0, SymPy 1.14.0 \\
 & Python 3.12.3, separate CPU host \\
\bottomrule
\end{tabular}
\end{table}

% ------------------------------------------------------------------------------
\section{Ceilings and Premiums}\label{sec:results}

\subsection{Per-model accuracy}
\begin{figure*}[t]\centering
\includegraphics[width=6.5in]{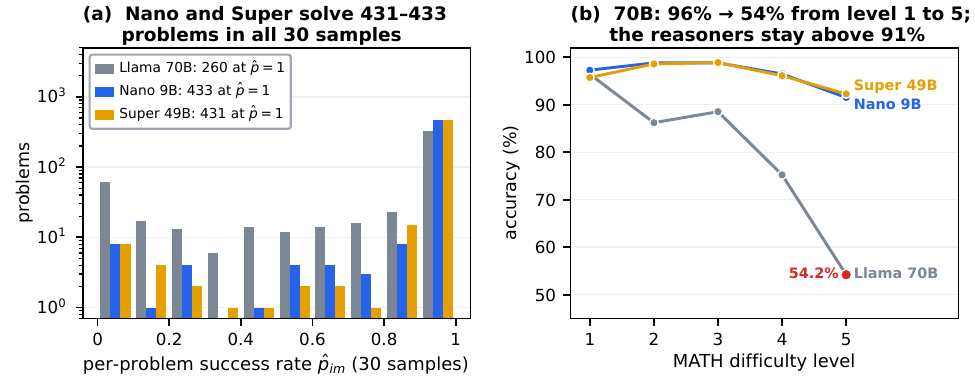}
\caption{Per-model results.  (a) Distribution of per-problem success rates $\hat p_{im}$ over
30 samples (log scale); the legend gives the number of problems each model solves in
every sample.  (b) Accuracy by MATH difficulty level.}
\label{fig:models}
\end{figure*}

Table~\ref{tab:models} and Fig.~\ref{fig:models}(a) summarize the three models. Both
reasoning models are near the top of the benchmark: Nano and Super each solve
\accNano\% of all samples (Table~\ref{tab:models}), and each answers more than $430$ problems correctly in
all $30$ samples. The 70B baseline, Super's parent, reaches \accSeventy\%. Super beats it on
\hSuperGtSeventy\ problems and loses on \hSeventyGtSuper; the two reasoning models
split their disagreements almost evenly (Nano higher on \hNanoGtSuper, Super higher on
\hSuperGtNano). No problem is unsolved by all three models, and
\hAllPerfect\ problems are solved by every model in every sample.

\begin{table*}[t]
\centering\footnotesize
\caption{Per-model results over $500\times30$ samples.}
\label{tab:models}
\begin{tabular}{@{}lrrrrrr@{}}
\toprule
Model & Acc.\ (\%) & All-30 & None-30 & Median & $p_{99}$ & Capped \\
 & $\bar{\hat p}$ & correct & correct & tokens & tokens & at $65{,}536$ \\
\midrule
Llama-3.3-70B & 76.19 & 260 & 45 & 463 & 2,039 & 6 \\
Nemotron Nano 9B v2 & 96.13 & 433 & 2 & 1,890 & 18,096 & 9 \\
Nemotron Super 49B v1.5 & 96.09 & 431 & 6 & 2,249 & 22,804 & 16 \\
\bottomrule
\end{tabular}

\end{table*}

Sampling noise is also model-dependent. The 70B model has \mixedSeventy\ problems
with $0<\hat p<1$ and produces \distinctSeventy\ distinct final answers per problem
on average; Nano and Super have \mixedNano\ and \mixedSuper\ such problems and
\distinctNano\ and \distinctSuper\ distinct answers. Single-sample labels are
therefore far less noisy for the reasoning models, which is consistent with the small
premiums below.

\begin{figure*}[t]\centering
\includegraphics[width=6.5in]{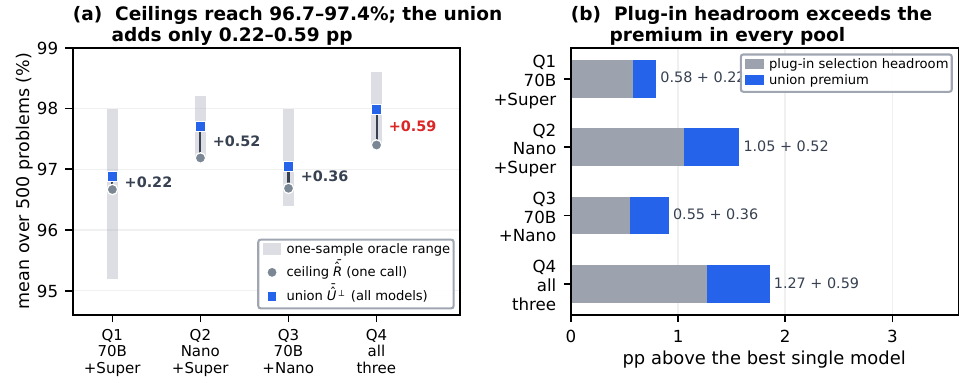}
\caption{Ceilings and premiums per pool (Q1 = 70B + Super, Q2 = Nano + Super, Q3 = 70B + Nano,
Q4 = all three).  (a) Single-commit ceiling $\bar{\hat R}$ (one call, clairvoyant), product
union $\bar{\hat U}^{\perp}$ (all models, gold-based selection) and the premium between
them; the grey band is the range, over the 30 sample indices, of an oracle computed from
one recorded sample per model.  (b) Gain over the best single model, split into
empirical plug-in selection headroom ($\bar{\hat R}$ minus the best model's accuracy) and union
premium ($\bar{\hat U}^{\perp}-\bar{\hat R}$), in percentage points.}
\label{fig:pools}
\end{figure*}

\subsection{Single-commit ceilings and union premiums}
Table~\ref{tab:pools} and Fig.~\ref{fig:pools}(a) give the main result. Every pool's
ceiling $\bar{\hat R}$ lies between \RQOne\% and \RQFour\%, and the product union adds
between \DQOne\ and \DQFour\,pp. Even the marginally identified upper end of the
premium, the largest union any coupling consistent with the measured marginals
could produce, is at most \DmaxQFour\,pp. The premium is positive on at most
\PosQFour\ of the $500$ problems.

\begin{table*}[t]
\centering\footnotesize
\caption{Single-commit ceiling, product-coupling union and premium per pool, with the
marginally identified premium interval of~\cite{chen2026whatoracle}. Means over 500
problems; pp = percentage points.}
\label{tab:pools}
\begin{tabular}{@{}llrrrrr@{}}
\toprule
Pool & Members & $k$ & $\bar{\hat R}$ (\%) & $\bar{\hat U}^{\perp}$ (\%) & $\bar{\hat\Delta}^{\perp}$ (pp) & Marginal interval (pp) \\
\midrule
Q1 & 70B + Super & 2 & 96.67 & 96.88 & 0.215 & $[0,\ 0.400]$ \\
Q2 & Nano + Super & 2 & 97.19 & 97.70 & 0.518 & $[0,\ 0.760]$ \\
Q3 & 70B + Nano & 2 & 96.69 & 97.05 & 0.360 & $[0,\ 0.653]$ \\
Q4 & 70B + Nano + Super & 3 & 97.40 & 97.99 & 0.588 & $[0,\ 0.827]$ \\
\bottomrule
\end{tabular}

\end{table*}

Holding the protocol fixed and changing only the pool, the premium ranges from
\DQOne\ to \DQFour\,pp, consistent with the companion's finding that pool composition
is a material conditioning choice~\cite{chen2026whatoracle}. The contrast with the
companion study is illustrative rather than controlled. On the same $500$ problems, its
11-model pool had a full-pool product premium of $3.54$\,pp and an identified interval
of $[0,4.79]$\,pp~\cite{chen2026whatoracle}; here the full-pool premium is \DQFour\,pp
and the interval $[0,\DmaxQFour]$\,pp, about one-sixth of both. Models, pool size,
temperature and length cap all differ, so this is not an effect of model strength; it
shows that the premium is not a fixed property of the benchmark. Pairing Super with its parent (Q1) leaves the smallest
premium, \DQOne\,pp: once Super is in the pool, calling the 70B model as well
almost never helps.

\begin{figure*}[t]\centering
\includegraphics[width=6.5in]{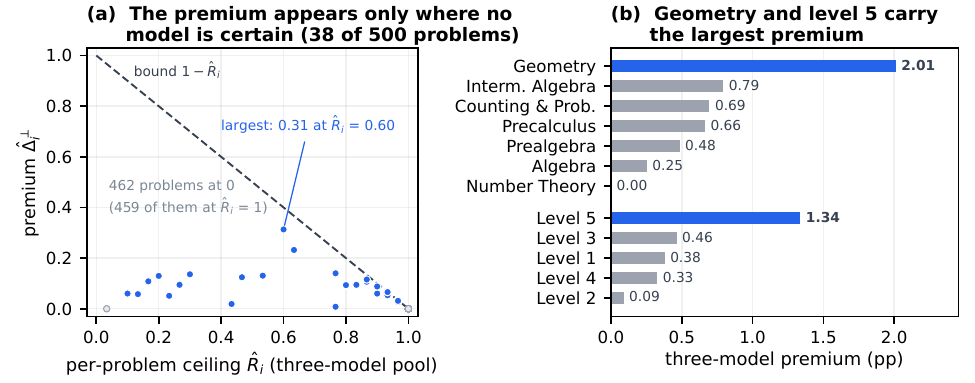}
\caption{Where the premium arises.  (a) Per-problem premium against ceiling for the
three-model pool; it is bounded by $1-\hat R_i$ and is zero whenever some model solves the
problem in every sample.  (b) Three-model premium by subject and by difficulty level.}
\label{fig:where}
\end{figure*}

Fig.~\ref{fig:where}(a) shows where the premium lives. It is non-zero only where no
model is certain, and is bounded by $1-\hat R_i$; with two models above $96\%$, few
problems leave room for a union to add anything.

\subsection{From best model to oracle}
Table~\ref{tab:headroom} and Fig.~\ref{fig:pools}(b) decompose the distance from the best single model to the
union. Always calling Nano yields \bestSingle\%. Committing to the model with the
highest estimated success on each problem gives the empirical ceiling of \RQFour\%, an
empirical plug-in selection headroom of \SelQFour\,pp; the product union adds
\DQFour\,pp more. This is the companion's decomposition~\cite{chen2026whatoracle}
instantiated with a baseline selected on the same data. The true ceiling bounds every
single-commit policy, but \SelQFour\,pp is a plug-in estimate from 30 samples per
cell, not a guaranteed router gain or an upper bound on one.

The last column and the grey bands of Fig.~\ref{fig:pools}(a) show how single-sample oracles can mislead. A benchmark that recorded
one sample per model and computed its oracle would report, for the three-model pool,
anything from \SdoMinQFour\% to \SdoMaxQFour\% depending only on which sample it kept
(mean \SdoMeanQFour\%, essentially $\bar{\hat U}^{\perp}$). For the 70B--Super pair the
range is \SdoMinQOne--\SdoMaxQOne\%, a swing of nearly $3$\,pp that dwarfs both the
selection headroom (\SelQOne\,pp) and the premium (\DQOne\,pp) it would be read as.

\begin{figure*}[t]\centering
\includegraphics[width=6.5in]{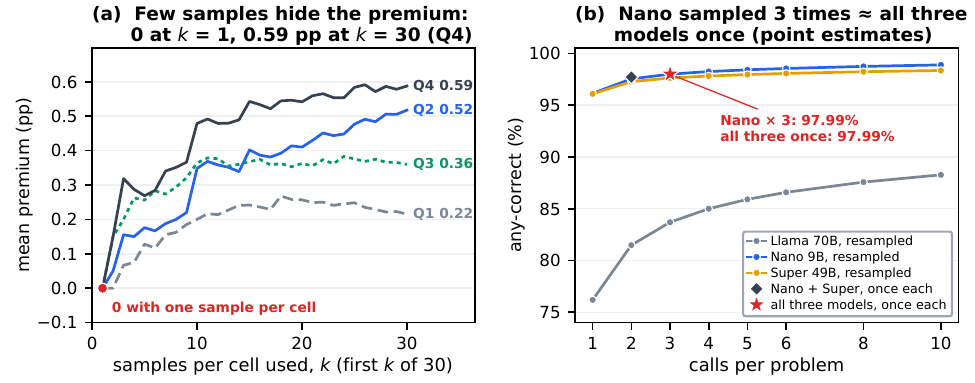}
\caption{Sampling.  (a) Mean plug-in premium recomputed from the first $k$ of the 30 frozen
samples per cell; with one sample every $\hat p$ is 0 or 1 and the plug-in premium is
identically zero.  (b) Any-correct rate against calls per problem: resampling one model
(lines, unbiased pass@$k$) versus calling different models once (markers, product
union); gold-based selection in both.}
\label{fig:sampling}
\end{figure*}

\subsection{Finite-sample behavior}
The premium estimate depends on how many samples are used. Fig.~\ref{fig:sampling}(a)
recomputes it from the first $k$ of the $30$ frozen samples. With one sample per cell
every $\hat p$ is $0$ or $1$, so the plug-in premium is identically
$\DQfourKone$; it rises to \DQfourKten\,pp at $k=10$ and \DQfourKthirty\,pp at $k=30$
for the three-model pool. A rising prefix path is not evidence of convergence. The
companion shows that, under iid sampling within cells and independent streams across
models, the plug-in premium has non-positive bias in expectation~\cite{chen2026whatoracle};
drawing every response with its own seed brings the present design closer to those
assumptions without proving them. A single-sample oracle, by contrast, can land above
or below the $30$-sample union depending on which sample is kept. Both are reasons to
report $k$ alongside any oracle number.

\subsection{Where the premium arises}
Table~\ref{tab:subject}, Fig.~\ref{fig:models}(b) and Fig.~\ref{fig:where}(b) break the results down by subject and
difficulty. The 70B model falls from \accLevelOneSeventy\% on level-1 problems to
\accLevelFiveSeventy\% on level 5, while Nano and Super stay above $91\%$
(\accLevelFiveNano\% and \accLevelFiveSuper\% on level 5). The premium concentrates on
the hardest level (\DlevelFive\,pp) and on geometry (\DGeometry\,pp), the subject on
which all three models are weakest; it is zero on number theory.

\subsection{Paired comparisons}
Table~\ref{tab:tests} reports exploratory two-sided Wilcoxon signed-rank tests on the
per-problem differences in success counts and a Friedman test across the three models,
over these $500$ fixed problems. Super and Nano each exceed the 70B model by about
$20$\,pp ($p<10^{-15}$). The Nano--Super difference of \wilNanoSuper\,pp is not
significant; this does not establish that the two models are equivalent.

% ------------------------------------------------------------------------------
\section{Resample or Reroute}\label{sec:resample}
The union compares calling several models; an obvious alternative at the same number
of calls is to sample the same model several times~\cite{brown2024monkeys,chen2026ror}.
Table~\ref{tab:passk} and Fig.~\ref{fig:sampling}(b) put the two side by side, with
gold-based any-correct selection in both. Resampling Nano three times reaches
\passThreeNano\%, practically the same as calling all three models once (\UQFour\%; no
equivalence test was run); with two calls, Nano twice (\passTwoNano\%) is within
$0.2$\,pp of Nano plus Super (\UQTwo\%). At an equal number of calls, the excess of the
union over the ceiling can therefore be matched without model diversity, in line with the
companion's point that the union premium is not an equal-cost routing
gain~\cite{chen2026whatoracle}. Nano's formal run took \wallShareNanoSuper\% of Super's
wall time for the same $15{,}000$ requests (one batch each, including loading and the
long-response tail; Section~\ref{sec:runaway}), so resampling the small reasoning model
looks like the cheaper option on these problems; the cost of a three-call strategy was
not measured. Resampling cannot
rescue a weak model: the 70B model reaches only \passThirtySeventy\% even with all
$30$ samples.

% ------------------------------------------------------------------------------
\section{The Cost of the Decoding Budget}\label{sec:runaway}
\begin{figure*}[t]\centering
\includegraphics[width=6.5in]{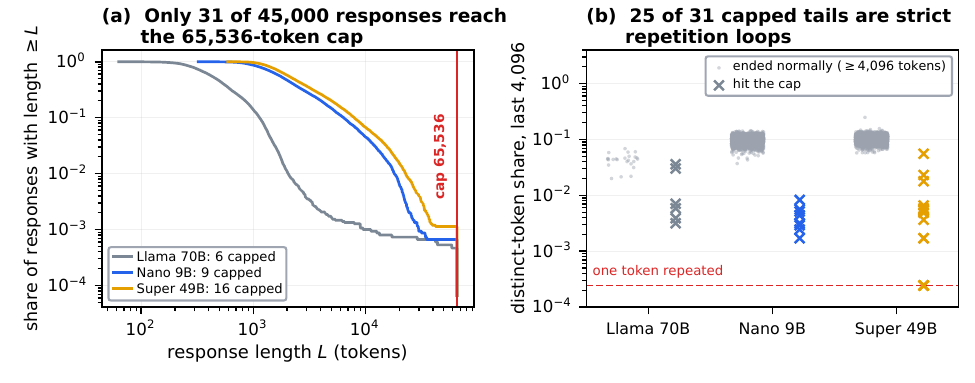}
\caption{Runaway generations.  (a) Share of responses at least $L$ tokens long; the red line
is the $65{,}536$-token cap.  (b) Share of distinct tokens in the last $4{,}096$ tokens of
responses that ended normally (grey, length $\ge4{,}096$) and of responses that hit the
cap ($\times$); the dashed line marks a single token repeated.}
\label{fig:loops}
\end{figure*}

\subsection{The cap rarely binds}
Fig.~\ref{fig:loops}(a) shows the length distributions. Ninety-nine percent of responses
end below \tokPninetynineSeventy, \tokPninetynineNano\ and \tokPninetynineSuper\
tokens for the 70B, Nano and Super models. Only \cappedTotal\ of $45{,}000$ responses
reached $65{,}536$ tokens (\cappedSeventy, \cappedNano\ and \cappedSuper).

To check whether the cap cut off genuine reasoning, we inspected only token IDs, after
generation and without gold. In \periodicTotal\ of the \cappedTotal\ capped responses
the last $4{,}096$ tokens repeat exactly with a period of at most $2{,}048$ tokens
(Table~\ref{tab:runaway}; the test required $95\%$ of positions to match, and all
\periodicTotal\ match on every position), four of them a single token repeated; this
describes the tail, not the whole response.
All capped responses have a tiny share of distinct tokens in their tail
(Fig.~\ref{fig:loops}(b)); for Nano and Super no response that ended normally has a tail
as repetitive as any capped one. The 70B separation is weaker, because five of its
few long normal responses also repeated before stopping. Truncating the existing outputs post hoc
(Table~\ref{tab:caps}) shows that a $32{,}768$-token cap would have cut
\lostThirtyTwoNano\ and \lostThirtyTwoSuper\ correct answers of Nano and Super and none
of the 70B model's; this re-scores existing outputs and is not a run with a smaller
cap, so it cannot show what a larger cap would have recovered.

\begin{figure*}[t]\centering
\includegraphics[width=6.5in]{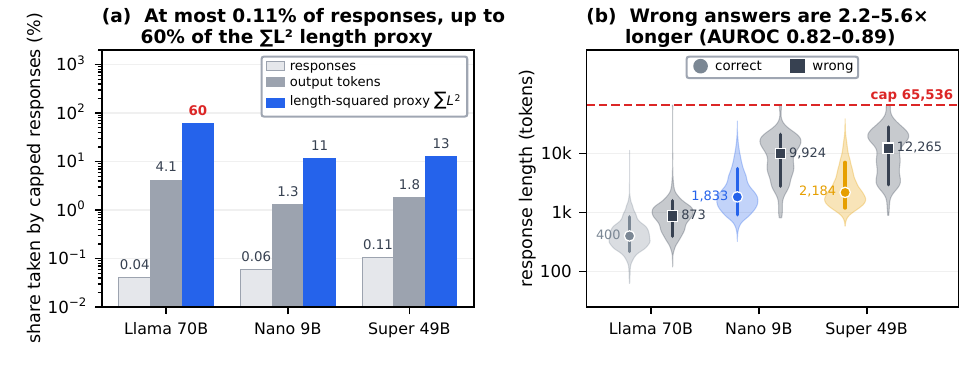}
\caption{Cost of the decoding budget.  (a) Share of responses, output tokens and the
output-length-squared proxy $\sum L^2$ taken by capped responses (log scale).  (b) Length
distributions of correct and wrong responses (shaded, log scale) with medians and
10--90\% ranges.}
\label{fig:cost}
\end{figure*}

\subsection{A few capped responses take an outsized share of a length-squared proxy}
In a dense Transformer each generated token attends to all earlier ones, so the
quadratic part of a response's attention cost grows with the square of its length. The
capped responses are at most $0.11\%$ of each model's responses but account for
\sqShareNano\%, \sqShareSuper\% and \sqShareSeventy\% of the output-length-squared proxy
$\sum_r L_r^2$ for Nano, Super and 70B (Fig.~\ref{fig:cost}(a)); the periodic-tail subset
alone accounts for \sqPeriodicNano\%, \sqPeriodicSuper\% and \sqPeriodicSeventy\%
(Table~\ref{tab:runaway}). The proxy is not measured GPU time, ignores length-linear
costs, and overstates the cost for Nano, a hybrid Mamba--Transformer whose Mamba layers
scale linearly with length. The engine's progress clock shows a matching concentration
in wall time: the 70B run completed its first $14{,}989$ requests in \headMinSeventy\
minutes and its last $11$ requests took a further \tailMinSeventy\ minutes,
\tailShareSeventy\% of its \wallFullSeventy-minute formal run including model
loading.

The fourth planned model points in the same direction. Llama-3.1-8B hit the cap in
\eightPreCapped\ of its \eightPreN\ preflight requests. In its formal run it finished
\eightEarly\ responses within twelve minutes and then only ${\sim}1.2$ requests per
minute: after \eightFinalHours\ hours, \eightFinal\ of $15{,}000$ were done. With
several more days in prospect, the run was stopped before any scoring and the model
removed (Section~\ref{sec:protocol}). Its formal responses were never scored or
inspected, so we report this as an observation consistent with the pattern above,
not as a measured loop rate.

\subsection{Execution cost and length as a signal}
Table~\ref{tab:exec} lists the wall-clock cost of each run. Nano generated
\tokTotalNano\,M tokens in \wallNano\ minutes and Super \tokTotalSuper\,M tokens in
\wallSuper\ minutes; at nearly equal observed accuracy, Nano needed
\gpuMinPerKNano\ GPU-minutes per $1{,}000$ correct answers against \gpuMinPerKSuper\ for
Super (formal wall time, including loading), a $\gpuRatioNanoSuper\times$ difference on
this GPU.

\begin{table}[t]
\centering\footnotesize\setlength{\tabcolsep}{3pt}
\caption{Formal-run execution on the B300 (one engine call per model, 15{,}000 requests).
Throughput includes model loading and the long-response tail.}
\label{tab:exec}
\begin{tabular}{@{}lrrrr@{}}
\toprule
Model & Wall (min) & Output tokens & Tokens/s & Capped \\
\midrule
Llama-3.3-70B & \wallSeventy & \tokTotalSeventy\,M & \tpsSeventy & \cappedSeventy \\
Nemotron Nano 9B v2 & \wallNano & \tokTotalNano\,M & \tpsNano & \cappedNano \\
Nemotron Super 49B v1.5 & \wallSuper & \tokTotalSuper\,M & \tpsSuper & \cappedSuper \\
\bottomrule
\end{tabular}
\end{table}

Wrong answers are also much longer than correct ones (Table~\ref{tab:lengths},
Fig.~\ref{fig:cost}(b)). The median wrong answer of Nano is \medWrongNano\ tokens
against \medCorrectNano\ for a correct one, and of Super \medWrongSuper\ against
\medCorrectSuper; wrong answers are $3.9\%$ of the reasoning models' responses but
about $15\%$ of their tokens. Length alone separates wrong from correct responses
with an AUROC of \aurocNano\ (Nano) and \aurocSuper\ (Super). This is a post hoc
description, not an evaluated early-stopping rule, but it suggests where budget-aware
decoding could save compute.

% ------------------------------------------------------------------------------
\section{Discussion}\label{sec:discussion}

\textbf{Relation to the companion study.} The new data support two of the companion's
empirical premises: pool composition alone moves the premium under one protocol, and at
an equal number of calls resampling one strong model practically matches the
multi-model union (as point estimates). The
remaining quantities instantiate its definitions on a new pool. Relative to the
companion's 11-model pool (3.54\,pp), the present premium is about one-sixth as large,
but models, pool size and decoding protocol changed together, so the contrast shows that
an oracle premium is a property of a pool and protocol rather than of the benchmark; it
does not isolate an effect of model strength. Because each response was drawn with its
own seed, the product coupling is closer to a design property than in the archived
setting; the same-index union differs from the product value by less than
$0.01$\,pp, which is compatible with, but does not establish, cross-model independence.

\textbf{Implications for routing evaluation.} An oracle gap is not evidence of routing
headroom unless it is reported together with the pool, the decoding protocol, the number
of samples per cell, and its split into selection headroom and union premium. On
MATH-500 with these models, the empirical plug-in selection headroom over a single
strong model is \SelQFour\,pp, an in-sample quantity rather than a bound on a learned
router; a small headroom in accuracy can still coexist with useful cost savings, which
this study did not evaluate.

\textbf{Implications for deployment.} Two observations are relevant to anyone serving
reasoning models. First, NVIDIA's NAS-derived Super 49B is about $20$\,pp more accurate
than the 70B model it was derived from, and the 9B Nano reaches Super's observed accuracy
in \wallShareNanoSuper\% of Super's formal-run wall time; on this benchmark, resampling a
small strong reasoner reaches practically the same empirical any-correct rate as a
three-model ensemble, with gold-based selection in both. Second, a handful of capped
generations take \sqShareNano--\sqShareSeventy\% of the length-squared proxy and, for the 70B run, stretch the end of the batch;
whether detecting them early would save time without losing answers was not evaluated.

% ------------------------------------------------------------------------------
\section{Limitations}\label{sec:limitations}
The results describe three models, one benchmark near its
ceiling, one decoding protocol and one GPU. MATH-500 leaves little room at the top;
the companion study found the largest premium on GPQA-Diamond~\cite{chen2026whatoracle},
which we did not run. The comparison with the 11-model pool spans different models,
temperatures and caps, so it cannot isolate model strength. The union and resampling
rates use gold-based selection and are not deployable selectors. The removal of
Llama-3.1-8B is a deviation from the original plan, made for cost before scoring. Our
loop diagnostic and cost proxy are post hoc and descriptive; the length-squared proxy
assumes dense attention, and throughput includes loading and tail effects of a single
large engine call. Only one scorer channel was used, so robustness to the scorer is not
tested, and the gold bundle is identified by hashes rather than included in the archive.

% ------------------------------------------------------------------------------
\section{Conclusion}\label{sec:conclusion}
Under a protocol fixed before generation, $45{,}000$ samples from three current models
on MATH-500 show that the empirical premium a hindsight oracle adds over the empirical
single-commit ceiling is at most \DQFour\,pp with strong reasoning models in the pool,
against $3.54$\,pp for the companion's small-model pool under a different protocol; that
an oracle built from one sample per model can differ from the $30$-sample union by more
than the premium itself; and that resampling the smallest strong model gives practically
the same empirical any-correct rate as a three-model union. The uniform $65{,}536$-token
cap was reached by \cappedTotal\ responses, most with periodic tails, yet these few
responses take \sqShareNano--\sqShareSeventy\% of a length-squared proxy of decoding cost. Routing papers should report
the ceiling, the premium, the samples per cell and the decoding budget together; the
execution record is archived to make that possible.

\section*{Artifact Statement}
The run record and all code are archived on Zenodo (DOI~\zenododoi).  The archive holds
the dated pre-commitments (v1, v2 and v4) and the scoring amendment, the seed manifests,
the append-only attempt ledger with per-file hashes, the engine-probe and preflight
receipts, all $45{,}000$ scored responses with their scorer statuses, the quarantined
Llama-3.1-8B attempt, the platform record, the generation, scoring and analysis code with
its tests, and the source of this report; one script regenerates every number, table and
figure from the archived record.  Two records are missing: the gold bundle, which the
scorer receipt pins by hash together with the dataset revision and can be rebuilt from
it, and the body of the superseded v3 commitment, which the v4 commitment identifies by
hash.  Host IP addresses and hardware serial numbers were removed from operational logs
before release, and every hash-bound file is byte-identical to the original.  The code is
released under the MIT License and the run record under CC~BY~4.0; model responses
remain subject to the licences of the models that produced them (Llama~3.3 Community
License; NVIDIA Open Model License).

\section*{Disclosures}
During the preparation of this work, the author used OpenAI Codex and Anthropic Claude
to assist with code implementation, experiment execution and monitoring, adversarial
checks, LaTeX drafting, and language editing.  The author independently determined
the research questions, protocol, estimands and final claims; reviewed and edited all
assisted output; and takes full responsibility for the code, data, figures, analysis,
and contents of the article.  All reported numerical results were produced by
recorded generation, scoring and analysis code over frozen run artifacts rather than
by generative text output.

The author is the founder of Krixvon; no Krixvon product, data, funding, or
other resource was used in this research.

This research received no specific grant from funding agencies in the public,
commercial, or not-for-profit sectors.

\bibliographystyle{IEEEtran}
\bibliography{refs}

% ---------------------------------------------------------------------------
% Supplementary Material (appended so that the arXiv record is one document).
% Sections and tables restart with the prefix S.
% ---------------------------------------------------------------------------
\setcounter{section}{0}\renewcommand{\thesection}{S\arabic{section}}
\setcounter{table}{0}\renewcommand{\thetable}{S\arabic{table}}
\setcounter{figure}{0}\renewcommand{\thefigure}{S\arabic{figure}}

\section*{Supplementary Material}
\section{Scorer outcomes}
Table~\ref{tab:status} lists every scorer status.  Responses with a verifier-handled
exception were all scored $0$ by the unchanged rule and are reported separately; no
pre-specified deterministic scorer failure occurred.

\begin{table}[H]
\centering\footnotesize
\caption{Scorer status counts per model (15{,}000 responses each).}
\label{tab:status}
\begin{tabular}{@{}lrrr@{}}
\toprule
Scorer status & 70B & Nano & Super \\
\midrule
correct & 11429 & 14420 & 14413 \\
not equivalent & 3507 & 556 & 526 \\
length-truncated & 6 & 9 & 16 \\
reasoning unterminated & 0 & 0 & 13 \\
reasoning malformed & 0 & 0 & 1 \\
no boxed answer & 0 & 2 & 0 \\
boxed empty & 0 & 0 & 1 \\
boxed unbalanced & 1 & 0 & 0 \\
unparseable (handled exc.) & 38 & 11 & 30 \\
not equivalent (handled exc.) & 19 & 2 & 0 \\
\midrule
total & 15000 & 15000 & 15000 \\
\bottomrule
\end{tabular}

\end{table}

\section{Additional tables for Sections V and VI}
Tables~\ref{tab:headroom}--\ref{tab:passk} give the decomposition plotted in
Fig.~\ref{fig:pools}, the per-subject and per-level results, the paired tests, and the
resample-versus-reroute comparison.

\begin{table}[H]
\centering\footnotesize\setlength{\tabcolsep}{3pt}
\caption{From the best single model to the union, per pool (Q1 = 70B + Super, Q2 = Nano + Super,
Q3 = 70B + Nano, Q4 = all three).  Headroom $=\bar{\hat R}$ minus the best single model's
accuracy; premium $=\bar{\hat U}^{\perp}-\bar{\hat R}$; the one-sample oracle range is the
minimum and maximum, over the 30 sample indices, of an oracle computed from one recorded
sample per model.}
\label{tab:headroom}
\begin{tabular}{@{}lrrrrrr@{}}
\toprule
Pool & Best & $\bar{\hat R}$ & $\bar{\hat U}^{\perp}$ & Headroom & Premium & One-sample \\
 & (\%) & (\%) & (\%) & (pp) & (pp) & oracle (\%) \\
\midrule
Q1 & 96.09 & 96.67 & 96.88 & 0.58 & 0.215 & 95.2--98.0 \\
Q2 & 96.13 & 97.19 & 97.70 & 1.05 & 0.518 & 97.2--98.2 \\
Q3 & 96.13 & 96.69 & 97.05 & 0.55 & 0.360 & 96.4--98.0 \\
Q4 & 96.13 & 97.40 & 97.99 & 1.27 & 0.588 & 97.4--98.6 \\
\bottomrule
\end{tabular}

\end{table}

\begin{table}[H]
\centering\footnotesize\setlength{\tabcolsep}{3pt}
\caption{Accuracy (\%) per model and three-model premium
$\bar{\hat\Delta}^{\perp}_{Q4}$ (pp) by subject and difficulty level.}
\label{tab:subject}
\begin{tabular}{@{}lrrrrr@{}}
\toprule
Subject / level & $n$ & 70B & Nano & Super & Premium \\
\midrule
Algebra & 124 & 93.6 & 99.5 & 99.4 & 0.25 \\
Counting \& Prob. & 38 & 74.4 & 95.7 & 97.6 & 0.69 \\
Geometry & 41 & 54.0 & 85.9 & 85.8 & 2.01 \\
Interm.\ Algebra & 97 & 59.0 & 94.5 & 93.9 & 0.79 \\
Number Theory & 62 & 80.0 & 99.6 & 99.6 & 0.00 \\
Prealgebra & 82 & 87.9 & 95.0 & 94.8 & 0.48 \\
Precalculus & 56 & 63.5 & 97.0 & 97.1 & 0.66 \\
\midrule
Level 1 & 43 & 96.4 & 97.3 & 95.7 & 0.38 \\
Level 2 & 90 & 86.2 & 98.8 & 98.6 & 0.09 \\
Level 3 & 105 & 88.5 & 98.8 & 98.9 & 0.46 \\
Level 4 & 128 & 75.2 & 96.5 & 96.1 & 0.33 \\
Level 5 & 134 & 54.2 & 91.5 & 92.3 & 1.34 \\
\bottomrule
\end{tabular}

\end{table}

\begin{table}[H]
\centering\footnotesize\setlength{\tabcolsep}{3pt}
\caption{Exploratory two-sided Wilcoxon signed-rank tests on per-problem differences
in success counts (zero differences dropped, SciPy 1.13.1) and a Friedman test.}
\label{tab:tests}
\begin{tabular}{@{}lrrrr@{}}
\toprule
Comparison & Mean diff.\ (pp) & Non-zero & $W$ & $p$ \\
\midrule
Super $-$ 70B & +19.89 & 251 & 1,266.5 & $<10^{-15}$ \\
Nano $-$ Super & +0.05 & 87 & 1,714 & 0.392 \\
Nano $-$ 70B & +19.94 & 243 & 1,138.5 & $<10^{-15}$ \\
\midrule
Friedman, 3 models &  &  & $\chi^2=315.8$ & $<10^{-15}$ \\
\bottomrule
\end{tabular}

\end{table}

\begin{table}[H]
\centering\footnotesize
\caption{Resample versus reroute at equal calls (gold-based any-correct; unbiased
pass@$k$ for resampling, product union for rerouting).}
\label{tab:passk}
\begin{tabular}{@{}lllr@{}}
\toprule
Calls & Strategy & Type & Any-correct (\%) \\
\midrule
2 & Nano, 2 samples & resample & 97.54 \\
2 & Super, 2 samples & resample & 97.27 \\
2 & Nano + Super (Q2) & reroute & 97.70 \\
\midrule
3 & Nano, 3 samples & resample & 97.99 \\
3 & Super, 3 samples & resample & 97.62 \\
3 & 70B + Nano + Super (Q4) & reroute & 97.99 \\
\bottomrule
\end{tabular}

\end{table}

\section{Additional tables for Section VII}
Tables~\ref{tab:runaway}--\ref{tab:lengths} give the capped-response counts, the post hoc
tighter-cap analysis, and the length statistics.

\begin{table}[H]
\centering\footnotesize\setlength{\tabcolsep}{3pt}
\caption{Capped responses: count, exactly periodic tails, and their share of
responses, output tokens and the output-length-squared proxy $\sum L^2$; the last
column is the share taken by the periodic-tail subset alone.}
\label{tab:runaway}
\begin{tabular}{@{}lrrrrrr@{}}
\toprule
Model & Capped & Periodic & \multicolumn{3}{c}{Share taken by capped (\%)} & Periodic \\
\cmidrule(lr){4-6}
 & & tail & responses & tokens & $\sum L^2$ & $\sum L^2$ (\%) \\
\midrule
Llama 70B & 6 & 3 & 0.040 & 4.15 & 60.2 & 30.1 \\
Nano 9B & 9 & 8 & 0.060 & 1.28 & 11.4 & 10.2 \\
Super 49B & 16 & 14 & 0.107 & 1.83 & 12.8 & 11.2 \\
\bottomrule
\end{tabular}

\end{table}

\begin{table}[H]
\centering\footnotesize
\caption{Post hoc: correct answers lost and output tokens saved under tighter caps.}
\label{tab:caps}
\begin{tabular}{@{}lrrr@{}}
\toprule
Model & \multicolumn{3}{c}{Correct lost / tokens saved (\%)} \\
 & cap $8{,}192$ & cap $16{,}384$ & cap $32{,}768$ \\
\midrule
Llama 70B & 0.01 / 5.4 & 0.00 / 4.3 & 0.00 / 2.6 \\
Nano 9B & 4.64 / 11.6 & 0.51 / 2.8 & 0.01 / 0.7 \\
Super 49B & 7.33 / 16.4 & 1.19 / 5.3 & 0.03 / 1.0 \\
\bottomrule
\end{tabular}

\end{table}

\begin{table}[H]
\centering\footnotesize\setlength{\tabcolsep}{3pt}
\caption{Length of correct and wrong responses, share of wrong responses and of their output
tokens, AUROC of length as a wrong-answer signal, and GPU-minutes per 1{,}000 correct
answers.}
\label{tab:lengths}
\begin{tabular}{@{}lrrrrrr@{}}
\toprule
Model & \multicolumn{2}{c}{Median tokens} & \multicolumn{2}{c}{Wrong share (\%)} & AUROC & GPU-min \\
\cmidrule(lr){2-3}\cmidrule(lr){4-5}
 & correct & wrong & resp. & tokens & length & per 1k \\
\midrule
Llama 70B & 400 & 873 & 23.8 & 42.6 & 0.82 & 6.4 \\
Nano 9B & 1,833 & 9,924 & 3.9 & 14.7 & 0.89 & 5.4 \\
Super 49B & 2,184 & 12,265 & 3.9 & 15.1 & 0.88 & 25.5 \\
\bottomrule
\end{tabular}

\end{table}

\end{document}